\documentclass[11pt,a4paper]{article}
\usepackage{acl2015}
\usepackage{times}
\usepackage{url}
\usepackage{latexsym}

\usepackage{times}
\usepackage{graphicx}
\usepackage{latexsym}

\usepackage{amssymb}
\usepackage{graphicx}
\usepackage{amsmath}

\usepackage{times}
\usepackage{multirow}
\usepackage{threeparttable}
\usepackage{array}
\usepackage{url}
\usepackage{latexsym}
\usepackage{graphicx}
\usepackage{amssymb}
\usepackage{amsmath}
\usepackage{algorithm} %format of the algorithm
\usepackage{algorithmic} %format of the algorithm
\usepackage{multirow} %multirow for format of table
\usepackage{amsmath}
\usepackage{xcolor}

\newcommand{\PreserveBackslash}[1]{\let\temp=\\#1\let\\=\temp}
\newcolumntype{C}[1]{>{\PreserveBackslash\centering}p{#1}}
\newcolumntype{R}[1]{>{\PreserveBackslash\raggedleft}p{#1}}
\newcolumntype{L}[1]{>{\PreserveBackslash\raggedright}p{#1}}
\allowdisplaybreaks

\DeclareMathOperator*{\argmax}{argmax}
\DeclareMathOperator*{\argmin}{argmin}

\usepackage{url}

\title{Towards Label Imbalance in Multi-label Classification with Many Labels}

\author{Li~Li~~~Houfeng~Wang\\
Key Laboratory of Computational Linguistics(Peking University), Ministry of Education, China\\
\texttt{\{li.l, wanghf\}@pku.edu.cn}\\
}

\date{}

\begin{document}
\maketitle
\begin{abstract}
In multi-label classification, an instance may be associated with a set of labels simultaneously. Recently, the research on multi-label classification has largely shifted its focus to the other end of the spectrum where the number of labels is assumed to be extremely large. The existing works focus on how to design scalable algorithms that offer fast training procedures and have a small memory footprint. However they ignore and even compound another challenge - the label imbalance problem. To address this drawback, we propose a novel Representation-based Multi-label Learning with Sampling (RMLS) approach. To the best of our knowledge, we are the first to tackle the imbalance problem in multi-label classification with many labels. Our experimentations with real-world datasets demonstrate the effectiveness of the proposed approach.
\end{abstract}

\section{Introduction}
Multi-label classification is supervised learning, where an instance may be associated with multiple labels simultaneously. Multi-label classification attracted increasing attention from various domains~\cite{vens2008decision,nicolas2013multi,sun2014multi,my} in these years. Due to several motivating real-life applications, such as image/video annotation~\cite{weston2011wsabie,kong2012multi} and query/keyword suggestions~\cite{agrawal2013multi}, the recent research on multi-label classification has largely shifted its focus to the other end of the spectrum where the number of labels is assumed to be extremely large~\cite{chen2012feature,agrawal2013multi,bi2013efficient,lin2014multi}.

Multi-label classification with many labels encounters the scalability challenge: how to design scalable algorithms that offer fast training procedures and have a small memory footprint. The standard multi-label classification approaches are computationally infeasible, when the number of labels is extremely large. For example, the simplest standard multi-label classification approach Binary Relevance (BR)  is not applicable for a multi-label classification problem with $10^4$ labels. BR trains a classifier for each label so that it need train $10^4$ classifiers. The high training time complexity makes it computationally infeasible.  BR is not applicable, not to mention the more sophisticated and computationally demanding approaches. There exists some works for multi-label classification with many labels. The mainstream approaches are called Label Space Dimension Reduction (LSDR)~\cite{hsu2009multi,tai2012multilabel,chen2012feature,lin2014multi,bi2013efficient}. LSDR encodes the high-dimensional label vectors into low dimensional code vectors. Then predictive models are trained from instances to code vectors. To predict an unseen instance, a low-dimensional code vector is firstly obtained with the predictive models,  and then be decoded for the label vector. Besides LSDR, there are another approaches with different style, and we call them Representation-Based Learning (RBL)~\cite{weston2011wsabie,yu2014large,rai2015large} approaches.  RBL learns representations for the instances and labels, and produces the predictions with these representations.

%The compression of label space is loss compression, which may consider the label with very little positive examples as noisy and drop the tinformation about this label. To address the label imbalance problem, the approach shall pay more attentions to the labels with little positive examples, however this approach drop the information about this label and compound the label imbalance problem.
%
%Since spanning the original label space from the small subset of class labels will losses information. As the first type of approaches, this approach will compound the label imbalance problem.
%
%This approach ignores the label imbalance problem.

However the above-mentioned approaches ignore and even compound an important problem: the label imbalance problem. The label imbalance problem is that the irrelevant labels of an instance are much more than relevant labels, and that some labels are irrelevant to more instances than other labels.  As the papers~\cite{spyromitros2011dealing,charte2013first,zhangtowards} pointed out, the label imbalance problem exists in the standard multi-label classification, and harms the performance.  The label imbalance problem becomes more serious in multi-label classification with many labels. Because more labels are irrelevant to an instance when the number of labels is large.  To show the phenomenon, we can use the imbalance ratio defined in~\cite{zhangtowards} to evaluate the label imbalance degree.  For a label, the imbalance ratio is the ratio of the number of irrelevant instances to the number of relevant instances.
\begin{eqnarray}
ImR_j &=& \frac{num\_of\_irrelevant\_instances}{num\_of\_relevant\_instances} \nonumber \\
ImR &=& \frac{1}{m} \sum_{j=1}^{m} ImR_j \nonumber
\end{eqnarray}
where $ImR_j$ denotes the imbalance ratio for the $j$-th label, $ImR$ denotes the average of the imbalance ratios. The high imbalance ratio indicates the serious label imbalance problem. The $Enron$ dataset~\cite{goldstein2006annotating} has 45 labels and its average imbalance ratio is 3. 34. The $Eurlex\_desc$ dataset~\cite{mencia2008efficient}  has 3993 labels and its average imbalance ratio is 1,378.58, much larger than the $Enron$ dataset's. The label imbalance problem in the $Eurlex\_desc$ dataset is more serious than that in the $Enron$ dataset. Hence we need attach more importance to the label imbalance problem in multi-label classification with many labels. However the existing approaches ignore the label imbalance problem.  Even LSDR compounds this problem. The labels with very little relevant instances contain little information. So the lossy compression in LSDR  may consider these labels as noisy and drop information about them. 

%The loss information may be the information about the label with very little positive examples. Since the label with very little positive examples contains less information, to drop the information about the label with very little positive examples will drops less information. To drop the information about the label with little positive examples will compound the label imbalance problem.

To address this drawback, we propose a novel Representation-based Multi-label Learning with Sampling (RMLS) approach, which can tackles the label imbalance problem in multi-label classification with many labels. To the best of our knowledge, we are the first to tackle the imbalance problem in multi-label classification with many labels. RMLS is a RBL approach and employs a representation learning framework with a sampling strategy.

\section{Related Works}
\subsection{Multi-label Classification with Many Labels}

\begin{figure}[!h!t]
\centering
\includegraphics[width=3in,height=2in]{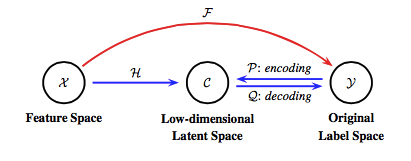}
\caption{An illustration of the principles of multi-label classification approaches (red) and those with LSDR (blue).}
\label{fig:lsdr}
\end{figure}

We categorize the existing approaches for multi-label classification with many labels into two types: Label Space Dimension Reduction (LSDR) and Representation-Based Learning (RBL). Figure~\ref{fig:lsdr} (it is from~\cite{lin2014multi}) is the schematic diagram  of LSDR.  LSDR encodes the high-dimensional label vectors into low dimensional code vectors. Then predictive models are trained from instances to code vectors. To predict an unseen instance, a low-dimensional code vector is firstly obtained with the learnt predictive models,  and then be decoded for the label vector.

Compressive Sensing (CS)~\cite{hsu2009multi} is the first LSDR approach. Specifically, CS linearly encodes the original label space as compressed sensing  and uses standard recovery algorithms for decoding.  Principle Label Space Transformation (PLST)~\cite{tai2012multilabel} performs PCA on the label matrix \(\pmb{Y}\) to get the compressing matrix $\pmb{V}$
\begin{eqnarray}
\pmb{V} ^{*} = \argmax_{\pmb{V}^{T}\pmb{V} =  \pmb{I}}  Tr(\pmb{V}^{T}\pmb{Y}^T\pmb{Y}\pmb{V})
\end{eqnarray}
Using the compressing matrix, we can obtain code vector \(\pmb{c} = \pmb{y}\pmb{V}\). CS and PLST aim to find the compressing matrix with high recoverability. However they don't consider the predictability of the code vector. With high predictability, it will is easy to train the model to predict the code vector. Conditional Principal Label Space Transformation (CPLST)~\cite{chen2012feature}  considers the predictability, and optimizes the following problem to get the compressing matrix.
 \begin{eqnarray}
\pmb{V} ^{*} = \argmax_{\pmb{V}^{T}\pmb{V} =  \pmb{I}}  Tr(\pmb{V}^{T}\pmb{Y}^T\pmb{X}
\pmb{X}^{+}\pmb{Y}\pmb{V})
\end{eqnarray}
CPLST argues that the compressing matrix obtained by this way can balances the predictability with recoverability. Feature-aware Implicit label space Encoding (FaIE)~\cite{lin2014multi} balances predictability with recoverability, and optimize the following problem.
\begin{eqnarray}
\pmb{V} ^{*} = \argmax_{\pmb{V}^{T}\pmb{V} =  \pmb{I}}  Tr(\pmb{V}^{T}(\pmb{Y}\pmb{Y}^{T}+\alpha \pmb{X}\pmb{X}^{+}))\pmb{V})
\end{eqnarray}
where $\alpha$ denotes a parameter specified by users. Column Subset Selection for Multi-Label (CSS$\_$ML)~\cite{bi2013efficient} seeks to select exactly $k$ representative labels so as to span all labels as much as possible. Then CSS$\_$ML learns $k$ classifiers for these selected labels. For unseen instance, CSS$\_$ML predicts $k$ selected labels and spans the predictions for all labels.  CSS$\_$ML can be considered as a special LSDR approach.

RBL learns representations for instances and labels, and produces the predictions with these representations. Web Scale Annotation by Image Embedding (WSABIE)~\cite{weston2011wsabie} trains the representation model by minimizing the Weighted Approximate-Rank Pairwise (WARP) loss function.  Low rank Empirical risk minimization for Multi-Label Learning (LEML) \cite{yu2014large} develops a fast optimization scheme for the representation model with different loss functions, and analyses the representation model's generalization error. Bayesian Multi-label Learning via Positive Labels (BMLPL)~\cite{rai2015large} uses the topic model to represent instance, and learns the model with only relevant labels

% \cite{yu2014large} formulates the multi-label classification as learning a low-rank linear model \(Z \in R^{d\times m}, rank(Z) <= k\). Since the rank is limited to be less than \(k\), the optimization problem becomes as follows.
%\begin{eqnarray}
%\hat{Z} &= & argmin_{Z} \sum_{i=1}^{n}\sum_{j=1}^{m} \ell(Y_{i,j},f^{j}(\pmb{x}_i;Z))+\lambda r(Z) \nonumber \\
%&s.t.& \quad Z = W^{T}H
%\end{eqnarray}
%where \(W \in R^{d\times k}, H \in R^{k \times m}\).

\subsection{Label Imbalance Problem}
The label imbalance problem has attracted some attention from the multi-label classification community. One solution to label-imbalance multi-label learning is to train a classifier for a label and deal with the skewness in each classifier via popular binary imbalance techniques such as random or synthetic undersampling/oversampling ~\cite{spyromitros2011dealing,tahir2012inverse,charte2013first,charte2015addressing}.  The paper~\cite{zhangtowards} improves this approach by aggregating one binary-class imbalance learner corresponding to the current label and several multi-class imbalance learners coupling with other labels for prediction. Besides integrating binary decomposition, Petterson et al~\cite{petterson2010reverse} and Dembczynski et al~\cite{dembczynski2013optimizing} address the label imbalance problem by directly optimizing imbalance-specific metric.

All of the above-mentioned approach solve the label imbalance problem by incorporating more correlations or designing more complex algorithms. These approaches are so complex that they are only applicable to the multi-label learning with the number of labels assumed to be small. In this paper, we aim to addressing the label imbalance problem in multi-label classification with many labels.

\section{Models}
\subsection{Preliminaries}

Let \(\mathcal{X}\) denote the instance feature space, and \(\mathcal{Y} = \{0,1\}^m\) denote label space with $m$ labels. A instance $\pmb{x} \in \mathcal{X} $ is associated with a label vector \(\pmb{y}=(y^1,y^2,...,y^m)\), where $y^j=1$ denotes the $j$-th label  is relevant to the instance and $y^j=0$ otherwise. The goal of multi-label learning is to learn a function \(\mathbf{f}:\mathcal{X} \rightarrow \mathcal{Y}\). In general, the function $\mathbf{f}$ consists of \(m\) functions, one for a label, i.e., \(\mathbf{f}(\pmb{x}) = [f^1(\pmb{x}), f^2(\pmb{x}),...,f^{m}(\pmb{x})]\), where $f^j(\pmb{x})$ is the prediction of the relevance between the instance $\pmb{x}$ and the $j$-th label.

\subsection{Representation Learning}

\begin{figure}
\centering
\includegraphics[width=3in,height=1.5in]{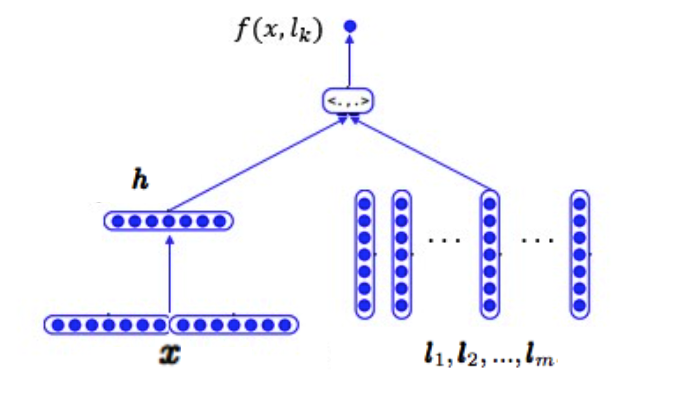}
\caption{An illustration of the representations learning framwork.}
\label{fig:structure}
\end{figure}

The architecture of RMLS is as shown in figure~\ref{fig:structure}. The feature vector \(\pmb{x}\) is mapped to a low-dimension feature representation vector \(\pmb{h}\) with a mapping matrix $\pmb{W}$
\begin{eqnarray}
\pmb{h} &=& \theta(\pmb{x} \pmb{W}) 
\end{eqnarray}
where $\theta$ is an activation function. Each label corresponds a low-dimension label representation vector, denoted by $\pmb{l}_1,\pmb{l}_2,...,\pmb{l}_m$. The dimension of the feature representation vector and the label representation vector are identical. The prediction for the $j$-th label, denoted by $f^j(\pmb{x})$, is produced by the inner dot of the feature representation vector \(\pmb{h}\) and the $j$-th label representation vector $\pmb{l_j}$. We add an activation function $\sigma$ to the inner dot, for example, the logistic function.

\begin{eqnarray}
f^j(\pmb{x}) &=& \sigma(\pmb{h}^T \pmb{l}_j)
\end{eqnarray}
where $\sigma$ is an activation function. our family of models have constrained norms:
\begin{eqnarray}
||W||_F^2 &\le& C_1 \nonumber \\
||l_j||_2^2 &\le& C_2, j=1,...,m 
\end{eqnarray}
The constrained norms acts as a regularizer in the same way as is used in lasso ~\cite{tibshirani1996regression}

%RMLS is motivated by feature-based matrix factorization~\cite{chen2011feature}. The feature-based matrix factorization is an abstract matrix factorization model that uses features to describe user/item factors. An instance in presentation learning framework can be consider as a user in feature-based matrix factorization, and its factor is the instance representation vector described by its features. A label in presentation learning framework can be consider as an item, and its factor is the label representation vector. 

If we set the dimension of the feature representation vector to be $k$, the number of parameters mapping an instance $\pmb{x}$ to an feature representation vector $\pmb{h}$ is $d\times k$, the number of parameters of all label representation vectors is $k \times m$. So the total number of parameters of RMLS is $d\times k + k \times m$. The simplest standard multi-label classification model BR trains $m$ classifiers. If the classifier in BR is a linear model with $d$ parameters, the total number of parameters of BR is $d \times m$.  Generally speaking, $k$ is much less than $min(d,m)$, so that the number of parameters of RMLS is much less than that of BR. The less parameters mean less training cost.

Both of LSDR and RBL reduce the number of parameters by this way. LSDR encodes the label vectors into the code vectors and generates a recovery matrix $\pmb{R}_{k \times m}$~\footnote{In this paragraph, we show the size of matrixes with subscripts}. Then LSDR learns regression models $\pmb{G}_{d\times k}$ mapping from the instances to the code vectors. The total number of parameters of LSDR is $d\times k + k \times m$. RBL learns a  model $\pmb{W}_{d \times k}$ mapping instances to low-dimension instance representation vectors. The number of parameters of label representation vectors is $k \times m$.  The total number of parameters of RBL is $d\times k + k \times  m$ too. The number of parameters of LSDR and RBL are the same, since the architecture of them are identical.  For an unseen instance $\pmb{x}$, LSDR produces the code vectors  with $\pmb{c} = \pmb{x}\pmb{G}_{d \times k}$, and then produces the prediction $\pmb{p}$ with $\pmb{p} = \pmb{c}\pmb{G}_{d \times k} = \pmb{x}\pmb{G}_{d \times k}\pmb{R}_{k \times m}$. RBL produces the instance representation vector with $\pmb{h} = \pmb{x}\pmb{W}_{d\times k}$. If we treat the label representation vectors as the columns of a label matrix $\pmb{L}_{k \times m}$,  and set the activation function to the linear function,  the prediction is produced with $\pmb{p} = \pmb{h}\pmb{L}_{k \times m} = \pmb{x}\pmb{W}_{d \times k}\pmb{L}_{k \times m}$. Hence the linear regression model $\pmb{G}_{d \times k}$ in LSDR is the equal of the mapping matrix $\pmb{W}_{d \times k}$ in RBL, and the recovery matrix $\pmb{R}_{k \times m}$ in LSDR is equal of the label matrix $\pmb{L}_{k \times m}$ in RBL. The difference between LSDR and RBL is how to obtain the parameters.  LDSR obtains parameters by linear algebra approaches, and RBL learns the model by gradient descent approaches. RMLS is not the first RBL approach, however we are the first to point out that LSDR and RBL have identical architectures and the same number of parameters. 

%However, the sof tmax training criterion adds additional computational overhead when performing
%mini-batch Stochastic Gradient Descent (SGD). Although we can use a plain SGD (i.e. mini-batch size is 1), mini-batch SGD is faster to converge and more stable. Assume the mini-batch size is m and the number of candidates is n, a total of m ¡Á n forward-backward passes over the network are performed to compute a similarity matrix (Fig. 3), while pairwise ranking criterion only needs 2¡Ám. We address this problem by grouping training pairs with same mention m into one minibatch
%{(d, ei)|ei ¡Ê C(m)}. Observe that if candidate entities overlap, they share the same forwardbackward
%path. Only m + n forward-backward passes are needed for each mini-batch now.
%
%Observe that , they share the same forward. Only m + n forward-backward passes are needed for each mini-batch now.

\subsection{Sampling Strategy}

Our goal is to identify relevant labels from irrelevant labels. To obtain this goal, we minimize a loss function over the training set to get the model parameters $\pmb{W}$ and $\pmb{l}_j$. The loss function $\mathcal{L}$ is as shown in the following formula. 
\begin{eqnarray}
\mathcal{L} = \sum_{j = 1}^{m}\ell(f^j(\pmb{x}),y^{j}) 
\end{eqnarray}
$\ell$ denotes the classification loss function. Different classification loss functions can be used as $\ell$, for example, cross entropy loss, least square loss and L2 hinge loss. 

With the serious label imbalance problem, the cost of classifying relevant labels as irrelevant is higher than that of classifying irrelevant labels as relevant.  Incorporating this consideration into the loss function, the loss function becomes as follows. 
\begin{eqnarray}
\mathcal{L} =  \sum_{j \in P}\ell(f^j(\pmb{x}),y^j)+\frac{1}{C}\sum_{j \in N}\ell(f^j(\pmb{x}),y^j)
\label{costloss}
\end{eqnarray} 
Where $\pmb{P} = \{j|y^j=1\}$ is the set of relevant labels and $\pmb{N} = \{j|y^j=0\}$ is the set of irrelevant labels. 

With the loss function, the overall risk we want to minimize is 
\begin{eqnarray}
R(\pmb{f})  =\int  \sum_{j \in P}\ell(f^j(\pmb{x}),y^j)+\frac{1}{C}\sum_{j \in N}\ell(f^j(\pmb{x}),y^j) d_{p(\pmb{x},\pmb{y})} 
\label{costrisk}
\end{eqnarray}
An unbiased estimator of this risk can be obtained by stochastically sampling $\frac{|\pmb{N}|}{C}$ irrelevant labels with the uniform distribution, and minimizing the loss function over the relevant labels and the chosen irrelevant labels.  Then the loss function becomes as follows.
\begin{eqnarray}
\mathcal{L} =  \sum_{j \in \pmb{P}}\ell(f^j(\pmb{x}),y^j) + \sum_{j \in \pmb{S}}\ell(f^j(\pmb{x}),y^j)
\label{sampled}
\end{eqnarray}
Where $\pmb{S}$ denotes the set of the chosen irrelevant labels.  Minimization the formula~\ref{sampled} approximates to obtain the minimizer of risk~\ref{costrisk}.

We think,  an instance with more relevant labels contains more information so that $C$ in the formula~\ref{costloss} should be less.  The number of an instance's relevant labels is denoted by $|\pmb{P}|$, and the number of an instance's irrelevant labels is denoted by $|\pmb{N}|$. We set $C = \frac{1}{\alpha}\frac{|\pmb{N}|}{|\pmb{P}|}$ and get the number of the chosen irrelevant labels $|\pmb{S}| =\alpha \times |\pmb{P}|$. We sample $\alpha \times |\pmb{P}|$ irrelevant labels with the uniform distribution, where $\alpha$ is the sampling coefficient.  The sampling coefficient $\alpha$ is an important parameter specified by the user, and we suggest to set it to be 5.

Our family of models have constrained norm so that the $\ell_2$ norm is added to the minimization objective. The final minimization problem becomes as follows.
\begin{eqnarray}
\pmb{W}, \pmb{l} &=& \argmin_{\pmb{W}, \pmb{l} }\{  \sum_{j \in \pmb{P}}\ell(f^j(\pmb{x}),y^j) + \sum_{j \in \pmb{S}}\ell(f^j(\pmb{x}),y^j)    \nonumber \\
                            &&+ \lambda ||\pmb{W}||_F^2+ \lambda \sum_{j=1}^{m}||\pmb{l}_j||_2^2\} \nonumber \\
 |\pmb{S}| &=&\alpha \times |\pmb{P}|
 \label{samplingobjective}
 \end{eqnarray}  
where $\lambda$ denotes the regularization coefficient.  
%Under-sampling is a popular approach in dealing with class-imbalance problems, which uses only a subset of the majority class and thus is very efficient. The traditional under-sampling approach samples some instances associated with the majority class, and construct a training set with these sampled instances and all instances with the minority class. In the multi-label classification setting, an instance may be associated with the majority label and the minority label simultaneously. Hence the traditional under-sampling is not applicable. We employ anther sampling scheme. Given a instance, the sampling scheme samples some absent labels, and constructs a training set with these sampled absent labels and all present labels. With these sampling scheme, the optimization problem can be formulated as follows.
%
%
%where $\pmb{p}_i$ is a set with all member labels associated to the $i$-th instance, and $\pmb{s}_i$ is the result labels of sampling for the $i$-th instance. For example, an instance has the first and third label, then $\pmb{p}=[1,3]$. The sampling scheme samples the fourth and seventh label, then $\pmb{s} = [4,7]$.

\subsection{Training Our Models}

The mini-batch Stochastic Gradient Descent (SGD) is performed to the above-mentioned optimization problem. We use the Adagrad~\cite{duchi2011adaptive} to adapt the learning rate. 

The sampling labels may be biased and unstable. The straightforward approach to this problem is to train different models with different sampling results and employ the ensemble strategy. However, it is very expensive to train different models for the large scale multi-label classification with many labels. A practical solution to this problem is to sample different labels in every batch of mini-batch SGD. Let B denote the index of the labelled training data  in a batch,  the pseudocode for training RLML with a batch of labelled data is given in Algorithm~\ref{alg1}.

\begin{algorithm} %Ëã·š¿ªÊŒ
\caption{Mini-batch SGD with  a batch of labelled data} %Ëã·šµÄÌâÄ¿
\label{alg1} %Ëã·šµÄ±êÇ©
\begin{algorithmic}[1] %ŽËŽŠµÄ[1]¿ØÖÆÒ»ÏÂËã·šÖÐµÄÃ¿ŸäÇ°Ãæ¶ŒÓÐ±êºÅ
\REQUIRE $(\pmb{x}_i,\pmb{y}_i)$ where $i \in B$ ,   $\alpha$ %ÊäÈëÌõŒþ(ŽËŽŠµÄREQUIRE Ä¬ÈÏ¹ØŒü×ÖÎªRequire£¬ÔÚÉÏÃæÒÑ×Ô¶šÒåÎªInput)
\FOR{$i \in B$} %start sampling
\STATE $\pmb{S}_i = \{\}$
\STATE count = $\alpha \times |\pmb{P}_i|$
\REPEAT
\STATE Pick a random irrelevant label $\overline{y} \in \pmb{N}_i $
\STATE $\pmb{S}_i = \pmb{S}_i \bigcup \{ \overline{y}\}$
\STATE count -= 1
\UNTIL count == 0
\ENDFOR
\STATE Make a gradient step to minimize Eq.(\ref{samplingobjective})
\end{algorithmic}
\end{algorithm}

\section{Experiments}
%We now present experimental results in order to assess our proposed algorithm in terms of accuracy and scalability. As we shall see, the results unambiguously demonstrate the superiority of our approach over other approaches.
%

\subsection{Datasets}

We perform experiments on four real world datasets. These datasets are available online \footnote{\url{http://mulan.sourceforge.net/datasets.html} and \url{http://mlkd.csd.auth.gr/multilabel.html} and \url{https://www.kaggle.com/c/lshtc/data}}. To reduce the time cost, we only use the accessible labelled training part of the $Wiki$ dataset and select the labels with at least 5 relevant instances. Table \ref{datasetstable} shows these multi-label datasets and associated statistics where \(n\) denotes the number of instances, \(d\) denotes the number of features, \(m\) denotes the number of labels. 
\begin{table}
\centering
\caption{Multi-label datasets and associated statistics.}
\begin{tabular}{C{1.6cm} C{1.6cm}    C{1.6cm} C{1.6cm}  }
  dataset  & \(n\) &\(d\) & \(m\)      \\
\hline
$Enron$        & 1702      & 1001  & 53           \\
$Delicious$   & 16105       & 500  & 983        \\
$Eurlex\_desc$     & 19348      & 5000  & 3993           \\
$Wiki$                 & 28596     & 23495   & 50341          \\
\hline
\end{tabular}
\label{datasetstable}
\end{table}

%%%%%%%%%%%%%%%Evaluation Metrics%%%%%%%%%%%%%%%%%%%%
\subsection{Evaluation Criteria}

Compared with the single-label classification, the multi-label setting introduces the additional degrees of freedom, so that we need various multi-label evaluation metrics. We use three common evaluation metrics in our experiments. Let $\pmb{p}$ denotes the prediction vector. The $Hamming loss$ is defined as the percentage of the wrong labels to the total number of labels. 
\begin{equation}
 Hamming loss =  \frac{1}{m}{| \pmb{p} \Delta \pmb{y}|}
\end{equation}
where \(\Delta\) denotes the symmetric difference of two sets, equivalent to XOR operator in Boolean logic.

Let \(p_i\) and \(r_i\) denote the precision and recall for the \(i\)-th instance, which means that \(p_i = \frac{| \pmb{p}_i \bigcap \pmb{y}_i|}{| \pmb{p}|}\) and that \(r_i = \frac{| \pmb{p} \bigcap \pmb{y}_i|}{| \pmb{y}_i|}\). The $F score$ is defined as follows. \begin{equation}
 F score = \frac{1}{n} \sum_{i=1}^{n}\frac{2p_ir_i}{p_i+r_i}
\end{equation}
The $F score$ is  a harmonic mean between precision and recall, and the higher F score means the better performance.

The $Accuracy$ in multi-label classification is the size of the intersection of predicted label set \(\pmb{p}\) and true label set \(\pmb{y}\) divided by the size of the union of this two set. The $Accuracy$ in multi-label classification is defined as follows:
\begin{equation}
  Accuracy = \frac{|\pmb{y} \cap \pmb{p}|}{|\pmb{y} \cup \pmb{p}|}
\end{equation}

%Let \(p_j\) and \(r_j\) denote the precision and recall for the \(j\)-th label. The label-based F score is a harmonic mean between precision and recall. Higher label-based F score means better performance.
%\begin{equation}
% F_{label} = \frac{1}{m} \sum_{j=1}^{m}\frac{2*p_j*r_j}{p_j+r_j}
%\end{equation}

%%%%%%%%%%%%%%%%approach setup%%%%%%%%%%%%%%%%%%%%
%\textbf{Competing approaches}
%A list containing details of the competing approach (including ours) is given below.
%
%\begin{itemize}
%
%\item[-]MLLF. Our proposed approach.
%
%\item[-]CPLST. CPLST is a LSDR approach proposed in~\cite{chen2012feature}. We use the code provided by the authors.
%
%\item[-]FaiE. FaiE is a LSDR approach proposed in~\cite{lin2014multi}.
%
%\item[-]ML$\_$CSSP. ML$\_$CSSP is a ML$\_$CSSP approach proposed in~\cite{bi2013efficient}.
%
%\item[-]LEML. LEML is a low-rank learning approach proposed in~\cite{yu2014large}. We use the code provided by the authors.
%\end{itemize}

%%%%%%%%%%%%%%% Result %%%%%%%%%%%%%%%%%%%%
\subsection{Experimentation Results}

\subsubsection{Performance Comparison}

We compare RMLS to some state-of-the-art appoaches and a baseline approach.
 
\begin{itemize}
\item[-] Principle Label Space Transformation (PLST)~\cite{tai2012multilabel}. PLST performs PCA on the label matrix to get the compressing matrix.
\item[-] Feature-aware Implicit label space Encoding (FaiE)~\cite{lin2014multi}. FaiE balances predictability with recoverability.
\item[-] Column Subset Selection for Multi-Label (CSS$\_$ML)~\cite{bi2013efficient}. CSS$\_$ML seeks to select exactly $k$ representative labels so as to span all labels as much as possible.
\item[-] Web Scale Annotation by Image Embedding (WSABIE)~\cite{weston2011wsabie}. WSABIE trains the representation model by  minimizing the Weighted Approximate-Rank Pairwise (WARP) loss function.
\item[-] Low rank Empirical risk minimization for Multi-Label Learning (LEML)~\cite{yu2014large}. LEML develops a fast optimization scheme for the representation model with different loss functions. 
\item[-] Baseline. The baseline classifies all labels as irrelevant labels. 
\end{itemize}

PLST, FaiE and ML$\_$CSSP are the LSDR approaches.  We use the open-source code $mlc\_lsdr$ \footnote{\url{https://github.com/hsuantien/mlc_lsdr}} for them. The  project  $mlc\_lsdr$ is developed by the author of PLST and CPLST.  In experiments, we use $mlc\_lsdr$'s default settings. WSABIE, LEML and our RMLS are the RBL approaches. We implement the code for WSABIE and LEML. When implementing LEML, we replace the gradient optimization scheme  by the least square minimization scheme. Since the least square minimization is more effective in the linear model. For our RMLS, we set the sampling coefficient $\alpha$ to 5, as suggested above. The regularization coefficients for WSABIE, LEML and our RMLS are set to 0.001. The dimension of the latent vectors $k$ (the dimension of the code vectors in LSDR and the dimension of the representation vectors in RBL)  is an important parameter.  We perform all algorithms on the $Enron$ dataset with $k = 25$ and $k=50$, and other datasets with $k=250$ and $k=500$.  The experiments are done in $five$-$fold$ cross validation. 

Table~\ref{effective1}, table~\ref{effective2} and table~\ref{effective3} show detail comparison results and we can draw two conclusions: 1) RMLS shows clear majorities of winning over the state-of-the-art approaches in terms of $F score$ and $Accuracy$, which demonstrates its effectiveness. In terms of $Hamming loss$, RMLS doesn't show superiorities. However, the winner in terms of $Hamming loss$ is the baseline, which predicts all labels as irrelevant labels. This implies that $Hamming loss$ is not a reasonable evaluation criteria for multi-label classification with the label imbalance, just like the  predictive accuracy isn't a good evaluation criteria for imbalance classification.  2) RBL approaches outperform LSDR approaches. The reason for it may be that the LSDR approaches make assumptions about compressing label space,  and that RBL approaches learn the label representations without making any assumptions.

\begin{table*}[!htb]
\small
\centering
\caption{Performance (mean\(\pm\)std.) of each approach in terms of $Hamming loss$.}
\begin{tabular}{|C{1.5cm}C{1.5cm} |C{2.5cm}C{2.5cm}C{2.5cm}C{2.5cm} |}
\hline
 Algorithm           &   k  &$Enron$ & $Delicious$   &$Eurlex\_desc$  &$Wiki$          \\
\hline

\multirow{2}*{RMLS}    & 25(0)    &$0.063\pm0.001\;\;$                              &$0.024\pm0.000\;\;$    &$0.002\pm0.000\;\;$  &$0.00049\;\;$\\
                                     & 50(0)   &$\textbf{0.055}\pm\textbf{0.001}\;\;$      &$0.024\pm0.000\;\;$   &$0.002\pm0.000\;\;$  &$0.00049\;\;$ \\
\hline

\multirow{2}*{PLST}     & 25(0)    &$0.082\pm0.003\;\;$     &$\textbf{0.018}\pm\textbf{0.000}\;\;$   &$0.002\pm0.000\;\;$ &- \\
                                     & 50(0)    &$0.090\pm0.003\;\;$    &$\textbf{0.018}\pm\textbf{0.000}\;\;$    &$0.003\pm0.000\;\;$  &-\\
\hline

\multirow{2}*{FaiE}      & 25(0)   &$0.082\pm0.003\;\;$     &$\textbf{0.018}\pm\textbf{0.000}\;\;$   &$0.002\pm0.000\;\;$    &- \\
                                    & 50(0)   &$0.091\pm0.003\;\;$     &$\textbf{0.018}\pm\textbf{0.000}\;\;$   &$0.003\pm0.000\;\;$  &- \\
\hline

\multirow{2}*{ML$\_$CSSP}    &  25(0)    &$0.079\pm0.001\;\;$    &$\textbf{0.019}\pm\textbf{0.000}\;\;$   &$0.002\pm0.000\;\;$ &$\textbf{0.00005}\;\;$\\
                                                &  50(0)    &$0.090\pm0.003\;\;$    &$\textbf{0.018}\pm\textbf{0.000}\;\;$   &$0.002\pm0.000\;\;$ &$0.00033\;\;$ \\
\hline

\multirow{2}*{WSABIE}    & 25(0)   &$0.070\pm0.008\;\;$    &$0.038\pm0.005\;\;$   &$0.015\pm0.000\;\;$  &$0.00146\;\;$\\
                                        &  50(0)  &$0.063\pm0.002\;\;$    &$0.091\pm0.012\;\;$   &$0.044\pm0.000\;\;$  & $0.00299\;\;$\\

\hline

\multirow{2}*{LEML}      & 25(0)   &$0.102\pm0.004\;\;$    &$0.026\pm0.001\;\;$     &$0.002\pm0.000\;\;$  &$0.00011\;\;$\\
                                      &  50(0)  &$0.103\pm0.004\;\;$    &$0.026\pm0.001\;\;$     &$0.003\pm0.000\;\;$  &$0.00011\;\;$ \\

\hline
Baseline                      & -   &$0.063\pm0.002\;\;$   &$\textbf{0.019}\pm\textbf{0.000}\;\;$   &$\textbf{0.001}\pm\textbf{0.000}\;\;$  &$\textbf{0.00005}\;\;$ \\

\hline
\end{tabular}
\label{effective1}
\end{table*}

\begin{table*}[!htb]
\small
\centering
\caption{Performance (mean\(\pm\)std.) of each approach in terms of $F score$.}
\begin{tabular}{|C{1.5cm}C{1.5cm} |C{2.5cm}C{2.5cm}C{2.5cm}C{2.5cm} |}
\hline
 Algorithm           &   k  &$Enron$ & $Delicious$   &$Eurlex\_desc$  &$Wiki$          \\
\hline

\multirow{2}*{RMLS}    &   25(0)   &$0.512\pm0.004\;\;$                              &$\textbf{0.329}\pm\textbf{0.007}\;\;$        &$\textbf{0.318}\pm\textbf{0.005}\;\;$    
                                     &$0.10209\;\;$\\
                                     &  50(0)   &$\textbf{0.587}\pm\textbf{0.011}\;\;$      &$\textbf{0.327}\pm\textbf{0.003}\;\;$        &$0.309\pm0.003\;\;$     
                                     &$\textbf{0.12154}\;\;$ \\
\hline

\multirow{2}*{PLST}     & 25(0)    &$0.450\pm0.011\;\;$      &$0.167\pm0.002\;\; $      &$0.201\pm0.003\;\;$  &- \\
                                     & 50(0)    &$0.442\pm0.011\;\;$      &$0.169\pm0.002\;\;$       &$0.230\pm0.004\;\;$  &-\\
\hline

\multirow{2}*{FaiE}       & 25(0)   &$0.451\pm0.011\;\;$       &$0.166\pm0.002\;\;$       &$0.201\pm0.003\;\;$ &- \\
                                     &  50(0)  &$0.442\pm0.011\;\;$       &$0.169\pm0.002\;\;$       &$0.230\pm0.004\;\;$  &- \\
\hline

\multirow{2}*{ML$\_$CSSP}    & 25(0)    &$0.421\pm0.006\;\;$     &$0.100\pm0.001\;\;$  &$0.085\pm0.010\;\;$   &$0.00039\;\;$ \\
                                                &  50(0)   &$0.442\pm0.011\;\;$      &$0.134\pm0.002\;\;$   &$0.135\pm0.010\;\;$  &$0.00043\;\;$\\
\hline

\multirow{2}*{WSABIE}    & 25(0)   &$0.436\pm0.052\;\;$    &$0.238\pm0.009\;\;$   &$0.131\pm0.002\;\;$  &$0.10928\;\;$ \\
                                        & 50(0)   &$0.498\pm0.022\;\;$    &$0.181\pm0.012\;\;$   &$0.052\pm0.000\;\;$  & $0.09631\;\;$\\

\hline

\multirow{2}*{LEML}     & 25(0)   &$0.429\pm0.007\;\;$     &$0.278\pm0.003\;\;$   &$0.246\pm0.005\;\;$  &$0.10505\;\;$ \\
                                     & 50(0)   &$0.424\pm0.007\;\;$     &$0.277\pm0.003\;\;$   &$0.255\pm0.005\;\;$  &$0.11022\;\;$ \\

\hline 

Baseline                      & -   &$0.000\pm0.000\;\;$   &$0.000\pm0.000\;\;$   &$0.000\pm0.000\;\;$  &$0.00000\;\;$ \\

\hline
\end{tabular}
\label{effective2}
\end{table*}

\begin{table*}[!htb]
\small
\centering
\caption{Performance (mean\(\pm\)std.) of each approach in terms of $Accuracy$.}
\begin{tabular}{|C{1.5cm}C{1.5cm} |C{2.5cm}C{2.5cm}C{2.5cm}C{2.5cm} |}
\hline
 Algorithm                      &   k  &$Enron$ & $Delicious$   &$Eurlex\_desc$  &$Wiki$          \\
\hline

\multirow{2}*{RMLS}      & 25(0)   &$0.380\pm0.007\;\;$                            &$\textbf{0.208}\pm\textbf{0.006}\;\;$   &$\textbf{0.207}\pm\textbf{0.004}\;\;$  
                                      &$0.06162\;\;$ \\
                                      & 50(0)   &$\textbf{0.456}\pm\textbf{0.013}\;\;$    &$\textbf{0.208}\pm\textbf{0.003}\;\;$   &\(0.196\pm0.002\;\;\)                       
                                      &$\textbf{0.07457}\;\;$ \\
\hline

\multirow{2}*{PLST}      & 25(0)    &$0.344\pm0.008\;\;$     &$0.107\pm0.001\;\;$  &$0.141\pm0.002\;\;$  &-\\
                                      & 50(0)    &$0.338\pm0.008\;\;$     &$0.109\pm0.001\;\;$  &$0.163\pm0.002\;\;$  &- \\
\hline

\multirow{2}*{FaiE}       & 25(0)   &$0.346\pm0.009\;\;$      &$0.107\pm0.001\;\;$   &$0.142\pm0.003\;\;$  &- \\
                                     &  50(0)  &$0.338\pm0.008\;\;$      &$0.110\pm0.001\;\;$   &$0.164\pm0.003\;\;$  &- \\
\hline

\multirow{2}*{ML$\_$CSSP}     & 25(0)    &$0.321\pm0.007\;\;$    &$0.062\pm0.008\;\;$   &$0.057\pm0.006\;\;$ &$0.00024\;\;$\\
                                                 & 50(0)    &$0.338\pm0.008\;\;$    &$0.086\pm0.001\;\;$   &$0.096\pm0.007\;\;$  &$0.00032\;\;$\\
\hline

\multirow{2}*{WSABIE}       & 25(0)    &$0.313\pm0.044\;\;$    &$0.141\pm0.006\;\;$   &$0.071\pm0.001\;\;$  &$0.06990\;\;$ \\
                                            & 50(0)   &$0.369\pm0.020\;\;$    &$0.102\pm0.008\;\;$   &$0.027\pm0.000\;\;$  & $0.06282\;\;$\\

\hline

\multirow{2}*{LEML}     & 25(0)    &$0.316\pm0.009\;\;$     &$0.179\pm0.003\;\;$    &$0.167\pm0.004\;\;$  &$0.07006\;\;$ \\
                                     & 50(0)    &$0.326\pm0.006\;\;$     &$0.178\pm0.002\;\;$     &$0.174\pm0.004\;\;$  &$\textbf{0.07381}\;\;$\\

\hline 

Baseline                      & -   &$0.000\pm0.000\;\;$   &$0.000\pm0.000\;\;$   &$0.000\pm0.000\;\;$  &$0.00000\;\;$\\

\hline
\end{tabular}
\label{effective3}
\end{table*}

\begin{table*}[!htb]
\small
\centering
\caption{The training time (seconds) of each approach.}
\begin{tabular}{|C{1.5cm}C{1.5cm} |C{2.5cm}C{2.5cm}C{2.5cm}C{2.5cm} |}
\hline
 Algorithm                      &   k  &$Enron$ & $Delicious$   &$Eurlex\_desc$  &$Wiki$          \\
\hline

\multirow{2}*{RMLS}      & 25(0)   &$9.85\pm1.24$        &$351.32\pm10.56$   &$1226.50\pm34.51$  
                                      &$9558.50\pm43.51$ \\
                                      & 50(0)   &$10.90\pm0.95$     &$575.45\pm16.40$    &$2436.35\pm58.84$                       
                                      &$15173.62\pm74.74$ \\
\hline

\multirow{2}*{PLST}      & 25(0)    &$0.39\pm0.02$       &$6.73\pm1.38$       &$220.14\pm10.34$  &-\\
                                      & 50(0)    &$0.41\pm0.05$      &$7.43\pm1.56$       &$246.95\pm11.84$  &- \\
\hline

\multirow{2}*{FaiE}       & 25(0)   &$1.34\pm0.11$      &$112.84\pm15.47$   &$408.18\pm23.14$  &- \\
                                     &  50(0)  &$1.51\pm0.21$     &$183.53\pm21.56$   &$567.92\pm31.97$  &- \\
\hline

\multirow{2}*{ML$\_$CSSP}     & 25(0)    &$0.28\pm0.01$    &$35.95\pm1.85$   &$330.87\pm30.51$ &$10693.83\pm134.04$\\
                                                 & 50(0)    &$0.53\pm0.03$    &$36.84\pm2.18$   &$334.89\pm32.14$  &$22207.16\pm189.30$\\
\hline

\multirow{2}*{WSABIE}       & 25(0)    &$10.75\pm2.35$    &$477.34\pm58.15$   &$767.40\pm81.36$  &$22974.65\pm203.71$ \\
                                            & 50(0)   &$11.78\pm3.51$    &$755.44\pm61.91$   &$1300.11\pm91.02$  & $41842.08\pm398.98$\\

\hline

\multirow{2}*{LEML}     & 25(0)    &$55.20\pm7.15$     &$39.34\pm4.51$    &$1231.34\pm18.95$  &$55847.23\pm481.56$ \\
                                     & 50(0)    &$55.34\pm8.01$     &$77.41\pm7.12$     &$1337.13\pm21.69$  &$57849.32\pm523.17$\\

\hline

\hline
\end{tabular}
\label{efficient}
\end{table*}

\begin{figure*}[!htb]
\centering
\includegraphics[width=6in,height=3in]{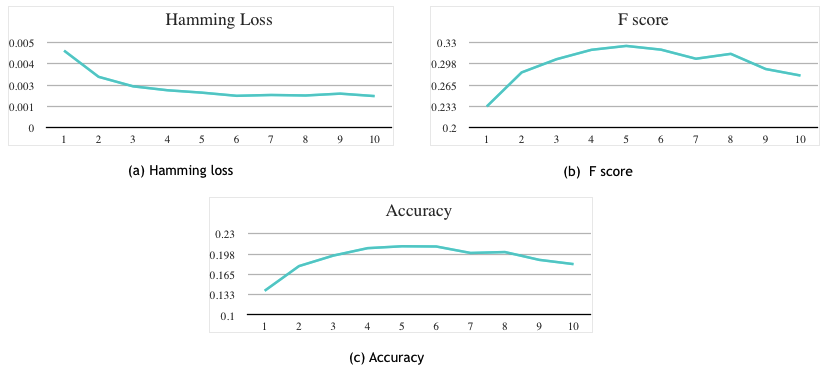}
\caption{Performance in terms of different evaluation criteria with different sampling coefficient $\alpha$.}
\label{fig:acc}
\end{figure*}

\subsubsection{Time Cost}

We also record the training time of each approach in table~\ref{efficient}.  We have some conclusions about the training time: 1) The training time of RBL approaches on small datasets ($Enron$, $Delicious$, $Eurlex\_desc$) are similar. But WSABIE and LEML spend much more time training on the $Wiki$ dataset than RMLS.  Because the sampling scheme in RMLS will reduce the time cost dramatically, when the number of labels is very large.  2) PLST and FaiE spend little time training on the small datasets ($Enron$, $Delicious$, $Eurlex\_desc$). But they run out of the memory and consume too much time on the $Wiki$ dataset, since both of them perform a partial SVD on the dense $50k\times50k$ matrix. 3) ML$\_$CSSP is the only LSDR approach that is applicable on the $Wiki$ dataset. But the performance is very poor.

\subsubsection{Influence of the Sampling Ratio}

%\begin{figure}
%\centering
%\includegraphics[width=3.5in,height=1.5in]{hamming.jpg}
%\caption{Performance in terms of $Hamming loss $ with different sampling coefficient $\alpha$.}
%\label{fig:hamming}
%\end{figure}
%
%\begin{figure}
%\centering
%\includegraphics[width=3.5in,height=1.5in]{fscore.jpg}
%\caption{Performance in terms of $F score $ with different sampling coefficient $\alpha$.}
%\label{fig:fscore}
%\end{figure}
%
%\begin{figure}
%\centering
%\includegraphics[width=3.5in,height=1.5in]{acc.jpg}
%\caption{Performance in terms of $Accuracy$  with different sampling coefficient $\alpha$.}
%\label{fig:acc}
%\end{figure}

To examine the influence of the sampling ratio , i.e., the parameter $\alpha$, we run RMLS with $\alpha$ varying from 1 to 10 with step size of 1. Due to the page limit, we only report results on the $Eurlex\_desc$ dataset, whereas experiments on other datasets get similar results. The detail results are shown in thefigure~\ref{fig:acc}

The $F score$ and $Accuracy$ are poor when the sampling ratio is small. As the sampling ratio grows large, these two evaluation criteria go up first and then down. When the sampling ratio is small, too many irrelevant labels are dropped, resulting the poor performance. When the sampling ratio is large, the number of irrelevant labels is much larger than the number of relevant labels, the label imbalance problem results in the poor performance. This implies that the sampling scheme with the proper sampling ratio $\alpha$ can handle the label imbalance problem and improve the performance.

The $Hamming loss$ goes down when the sampling ratio grows up.  When the sampling ratio is large, we achieve good performance in terms of $Hamming loss$ with the serious label imbalance problem. The reason for it may be that $Hamming loss$ is not a reasonable evaluation criteria, which has been uncovered in the performance comparison experiments.

\section{Conclusions}
In multi-label classification, an instance is associated with with a set of labels simultaneously. Recently, the researchers on multi-label classification focused on the multi-label learning with many labels. The existing approaches for multi-label learning with many labels ignore and even compound the label imbalance problem. To address this problem, we propose a novel Representation-based Multi-label Learning with Sampling (RMLS) approach. Our experimentations demonstrate the effectiveness of the proposed approach.% include your own 

\bibliographystyle{acl}
\bibliography{emnlp2015}

\end{document}